\documentclass[conference]{IEEEtran}
\IEEEoverridecommandlockouts

\usepackage{cite}
\usepackage{amsmath,amssymb,amsfonts}
\usepackage{algorithmic}
\usepackage{graphicx}
\usepackage{textcomp}
\usepackage{xcolor}
\usepackage{amssymb}
\usepackage{comment}
\usepackage{graphicx}
\usepackage{hyperref}
\usepackage{subcaption}
\usepackage{tikz}
\usepackage{amsmath}
\usepackage{multirow}
\usepackage{float} 
\usepackage{url}
\usepackage{underscore}
\usepackage{enumitem}
\usepackage{comment}
\def\BibTeX{{\rm B\kern-.05em{\sc i\kern-.025em b}\kern-.08em
    T\kern-.1667em\lower.7ex\hbox{E}\kern-.125emX}}
\begin{document}

\title{Advancing Ear Biometrics: Enhancing Accuracy and Robustness through Deep Learning}

\author{\IEEEauthorblockN{1\textsuperscript{st} Youssef Mohamed}
\IEEEauthorblockA{\textit{Computer science and information technology} \\
\textit{Egypt-Japan University of Science and Technology}\\
Alexandria, 21934, Egypt \\
youssef.khalil@ejust.edu.eg}
\and
\IEEEauthorblockN{2\textsuperscript{nd} Zeyad Youssef}
\IEEEauthorblockA{\textit{Computer science and information technology (E-JUST)} \\
\textit{Egypt-Japan University of Science and Technology (E-JUST)} \\
Alexandria, 21934, Egypt \\
zeyad.youssef@ejust.edu.eg}
\and
\IEEEauthorblockN{3\textsuperscript{rd} Ahmed Heakl}
\IEEEauthorblockA{\textit{Computer Science and Engineering} \\
\textit{Egypt-Japan University of Science and Technology (E-JUST)} \\
Alexandria, 21934, Egypt \\
ahmed.heakl@ejust.edu.eg}
\and
\IEEEauthorblockN{4\textsuperscript{th} Ahmed B. Zaky}
\IEEEauthorblockA{\textit{Computer science and information technology (E-JUST)} \\
\textit{Egypt-Japan University of Science and Technology}\\
Alexandria, 21934, Egypt \\
ahmed.zaky@feng.bu.edu.eg,ahmed.zaky@ejust.edu.eg}
}

\maketitle

\begin{abstract}

Biometric identification is a reliable method to verify individuals based on their unique physical or behavioral traits, offering a secure alternative to traditional methods like passwords or PINs. This study focuses on ear biometric identification, exploiting its distinctive features for enhanced accuracy, reliability, and usability.
While past studies typically investigate face recognition and fingerprint analysis, our research demonstrates the effectiveness of ear biometrics in overcoming limitations such as variations in facial expressions and lighting conditions.
We utilized two datasets: AMI (700 images from 100 individuals) and EarNV1.0 (28,412 images from 164 individuals). To improve the accuracy and robustness of our ear biometric identification system, we applied various techniques including data preprocessing and augmentation.
Our models achieved a testing accuracy of 99.35\% on the AMI Dataset and 98.1\% on the EarNV1.0 dataset, showcasing the effectiveness of our approach in precisely identifying individuals based on ear biometric characteristics.
\end{abstract}

\begin{IEEEkeywords}
Biometric identification, Ear biometrics, Data preprocessing, Convolutional neural networks
\end{IEEEkeywords}

\section{Introduction}
In recent years, the demand for secure automated identity systems has surged, driven by the imperative for robust security measures in various sectors. Consequently, there has been a notable uptick in the exploration and implementation of secure biometric systems characterized by exceptionally high classification accuracy \cite{1}. Much of the research in biometric identification has traditionally centered on modalities such as face recognition, gait analysis, palm-print identification, iris scanning, hand geometry, and EEG recognition. \newline
Face recognition utilizes facial features for identity verification, including facial landmarks, texture, and shape~\cite{2,3}. However, it is susceptible to variations caused by facial expressions and may struggle with low-light conditions.
Fingerprint analysis analyzes unique patterns in fingerprint ridges and valleys, and it is widely used due to high uniqueness and reliability~\cite{4,5}. Yet, it is prone to fingerprint spoofing attacks using artificial fingerprints and can be affected by skin damage or wear~\cite{6}.
Gait analysis identifies individuals based on their walking patterns and movements. It offers potential for continuous and passive authentication~\cite{7,8}. However. the accuracy can be affected by walking speed and may not be suitable for individuals with mobility issues.
Palm-print identification examines patterns in palm prints for identification. It can be used as a supplementary biometric modality~\cite{9,10}, but it is vulnerable to environmental factors such as dirt or moisture, which can affect accuracy.
Iris scanning captures the unique patterns in the iris. It provides high accuracy and resistance to forgery~\cite{11,12}, but with high cost and potential discomfort during scanning.
Hand-geometry analyzes the physical characteristics of the hand, including size and shape. It offers ease of use and quick enrollment~\cite{13,14}. Nevertheless, it requires physical contact, which may not be desirable in some scenarios and may not be suitable for individuals with hand deformities.
EEG recognition utilizes brainwave patterns for identity verification, including frequency, amplitude, and waveform characteristics. EEG signals can provide valuable insights into cognitive states and mental activities~\cite{15}. However, EEG signals are susceptible to artifacts caused by muscle movements, eye blinks, and environmental interference. Additionally, EEG data collection may require specialized equipment and electrodes, making it less suitable for certain real-world applications. \\
This study will discuss the application of ear biometric identification using preprocessing and deep-learning techniques. There is a growing recognition of the potential of ear biometrics as a reliable and effective means of identity verification, as it provides more reliable information for human identification, including its stability over time, ease of acquisition from a distance, uniqueness even for twins~\cite{16}, and the fact that they typically involve capturing less identifiable information compared to facial or fingerprint data, reducing the risk of privacy breaches. \\
Our work makes the following contributions:
\begin{itemize}
    \item This work advances ear biometric identification by introducing novel techniques, including zooming and augmentation, which significantly enhance model performance on both EarNV1.0 and AMI datasets. Specifically, our methodologies improve accuracy by 8\% on EarNV1.0 and 2\% on AMI, surpassing the base model. 
    \item Our approach yields a reliable performance boost of 2\% on AMI and 1-4\% on EarNV1.0 across various models, outperforming existing methods, including the state-of-the-art.
\end{itemize}

\section{Related Works}

Priyadharshini et al. \cite{17} conducted a study focusing on ear recognition utilizing CNNs. Their model, featuring a six-level architecture with batch normalization for stability, achieved impressive recognition rates of 97.36\% and 96.99\% on the IITD-II \cite{18} and AMI ear datasets, respectively. 
Hossain et al., 2021 \cite{19} explored ear biometric identification employing a combination of traditional ML techniques and deep learning-based models. Their approach integrated Independent Component Analysis (ICA) and Principal Component Analysis (PCA) for traditional ML, alongside the utilization of YOLOv3\cite{20}, a deep learning model, for image categorization and individual identification. The study utilized the EarVN1.0 Dataset, achieving a 90\% accuracy, while utilizing GPC resulted in 96\% accuracy.
\newline
Alkababji \& Mohammed, 2021 \cite{21} proposed a multi-stage ear localization and recognition system, leveraging Faster R-CNN\cite{22} for ear detection and localization, followed by feature extraction using AlexNet\cite{23}. Employing techniques such as PCA and genetic algorithms for feature reduction and selection, the system achieved promising results, with an ear detection accuracy of 97\%. 
\newline
Ajay et al. \cite{24} proposing an automated approach for ear recognition utilizing morphological operators and Fourier descriptors for ear segmentation, with feature extraction leveraging complex Gabor filters. Notably, log-Gabor filters exhibited superior performance, achieving high recognition accuracies of 96.27\% and 95.93\%, respectively, on databases comprising 125 and 221 subjects. The study highlights the importance of robust feature extraction techniques in ear recognition systems.
\newline
Banafshe et al. \cite{25} addressed challenges posed by partial occlusion in ear biometrics. Their novel approach utilized automatic image alignment and template construction, coupled with log-Gabor filters for feature extraction. Achieving a recognition rate of 97.4\%, the study demonstrates the effectiveness of their approach in handling occluded ear data, contributing to advancements in the field of ear biometrics.\\
Collectively, these studies signify the evolving landscape of ear recognition methodologies, showcasing the efficacy of both traditional ML techniques and deep learning approaches in addressing various challenges inherent in ear biometrics.

\section{Datasets}\label{sec:sample:sample1}

\subsection{AMI dataset}\label{sec:sample:ami}
\begin{figure}[H]
  \centering
  \captionsetup[subfigure]{justification=centering}
  \begin{subfigure}{0.18\textwidth}
    \includegraphics[width=\linewidth]{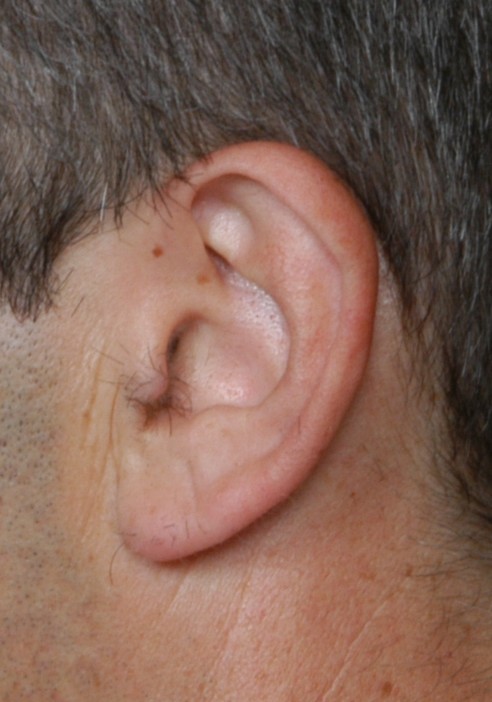}\par
    \caption{}
  \end{subfigure}
  \begin{subfigure}{0.18\textwidth}
    \includegraphics[width=\linewidth]{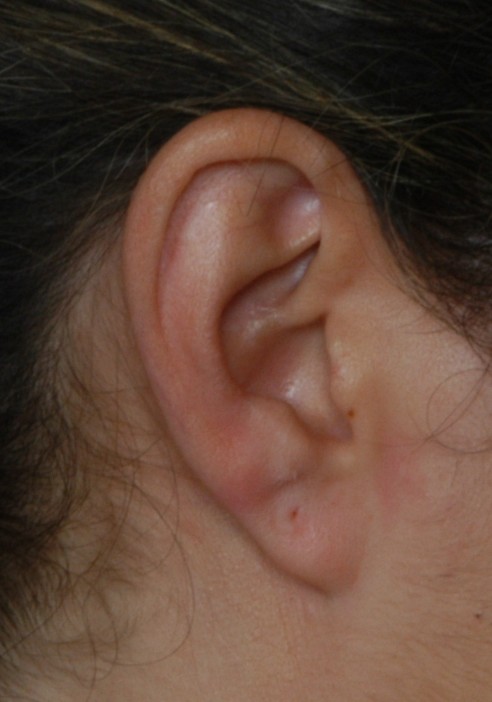}\par
    \caption{}
  \end{subfigure}
  \caption{Sample images from the AMI dataset.}
  \label{fig:five-images}
\end{figure}

The AMI \cite{26} ear dataset comprises 700 images representing 100 individuals, aged between 19 and 65 years. Each person is depicted by six images of their right ear and one image of their left ear. Among the right ear images, five showcase the individual looking right, left, up, down, and forward, while the sixth image is zoomed. All images maintain a resolution of 492x702 pixels. While the dataset's limitation lies in its relatively small image count, its advantages include a controlled environment and high resolution, facilitating easier model training. Figure ~\ref{fig:five-images} shows samples of different positions for images in the AMI dataset. 

\subsection{EarVN1.0 Dataset}\label{sec:sample:sample3}
\begin{figure}[H]
  \centering
  \begin{tabular}{c@{\hspace{0.2cm}}c@{\hspace{0.2cm}}c@{\hspace{0.2cm}}c@{\hspace{0.2cm}}c}
    \includegraphics[width=2cm]{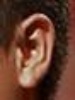} &
    \includegraphics[width=2cm]{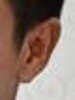} &
    \includegraphics[width=2cm]{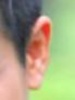} &
  \end{tabular}
  \caption{Samples from the EarNV1.0 dataset for the same person.}
  \label{fig:samples-NV}
\end{figure}

The EarVN1.0 dataset \cite{27}, developed in 2018, stands out as one of the most comprehensive publicly available collections of ear images. Comprising 28,412 images, it encompasses contributions from 98 male and 66 female individuals. These images were meticulously curated by isolating the ears from facial photographs, taking into account variations in location, size, and brightness. The disadvantage of this dataset is its low image resolution, while its advantage lies in the large number of images it contains. Figure~\ref{fig:samples-NV} shows samples of images for the same person in EarVN1.0 dataset.

Table \ref{tab:dataset-comparison} shows a comparison between the structure of AMI and EarNV1.0 datasets.
\begin{table}[H]
    \centering
    \begin{tabular}{ |p{3cm}|p{2cm}|p{2.4cm}|  }
        \hline
        Datasets & AMI & EarNV1.0 \\
        \hline
        \# Images & 700 & 28412 \\ \cline{1-3}
        Image resolution & $492 \times 702$ & Variable resolution  \\ \cline{1-3}
        \# individuals & 100 & 164 \\ \cline{1-3}
        \# images per person & 7 & 107 - 300 \\ \hline
    \end{tabular}
    \caption{Comparison between AMI and EarNV1.0 datasets}
    \label{tab:dataset-comparison}
\end{table}

\section{Data Preprocessing}\label{sec:sample:sample2}
Ear recognition systems require a meticulously prepared dataset to achieve accurate identification. The preprocessing stage, the initial hurdle, involves cleaning and transforming raw ear images to address challenges such as inconsistent lighting, head poses, noise, and image artifacts. Key preprocessing techniques for ear biometrics include zooming, ensuring consistent image size and specifying ear shape, contour detection, and isolating the ear from the background.

\subsection{Zooming}\label{sec:sample:sample}
Ear recognition relies on tiny differences in ear shapes to identify people. Zooming images in this system is crucial as it ensures that all images are of the same size. Additionally, zooming is performed to remove extraneous elements such as facial hair, ensuring that when the model is working, it focuses solely on the ear in the image.

\begin{figure}[htpb]
  \centering
  \begin{tabular}{cc} 
    \includegraphics[width=0.12\textwidth, height=0.12\textwidth]{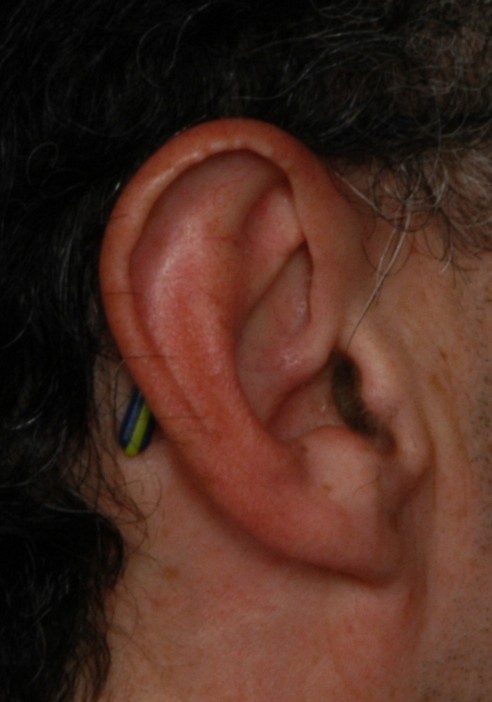} &
    \includegraphics[width=0.12\textwidth, height=0.12\textwidth]{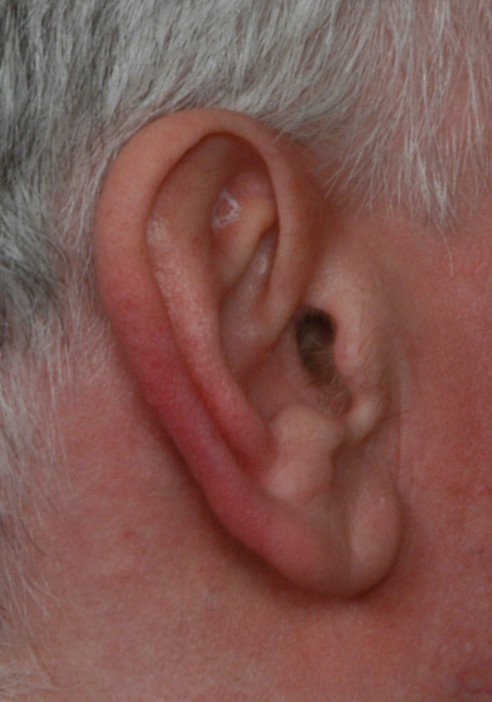} \\
    $\downarrow$ & $\downarrow$ \\
    \includegraphics[width=0.12\textwidth, height=0.12\textwidth]{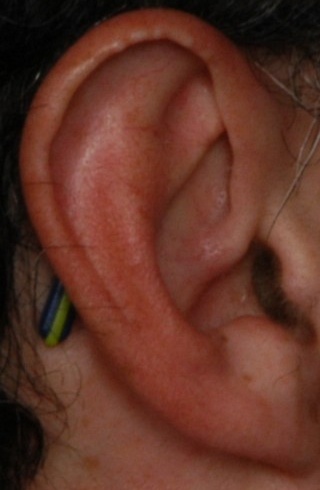} &
    \includegraphics[width=0.12\textwidth, height=0.12\textwidth]{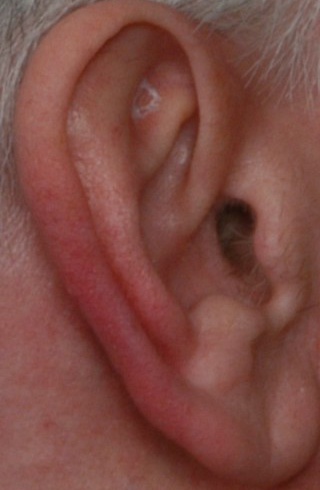} \\
  \end{tabular}
  \caption{Sample images before and after zooming.}
  \label{fig:four-AMI-images}
\end{figure}

For the AMI dataset, the images were zoomed from their original size of 492x702 pixels to 320x490 pixels. This output image size was determined through trial and error. Figure~\ref{fig:four-AMI-images} depicts two images before and after zooming.

For the EarNV1.0 dataset, zooming was not performed. This decision was made due to the variation in image resolutions and locations within the dataset, which makes it challenging to specify a uniform threshold for zooming across all images. Implementing zooming could result in over-zooming certain images, leading to the loss of important ear features such as edges and a decrease in the accuracy of the model. As it is shown in figure \ref{fig:samples-NV}, ear appears in different locations in images.

\subsection{Contour Detection}
\begin{figure}[H]
    \centering 
    \includegraphics[width=0.485\textwidth]{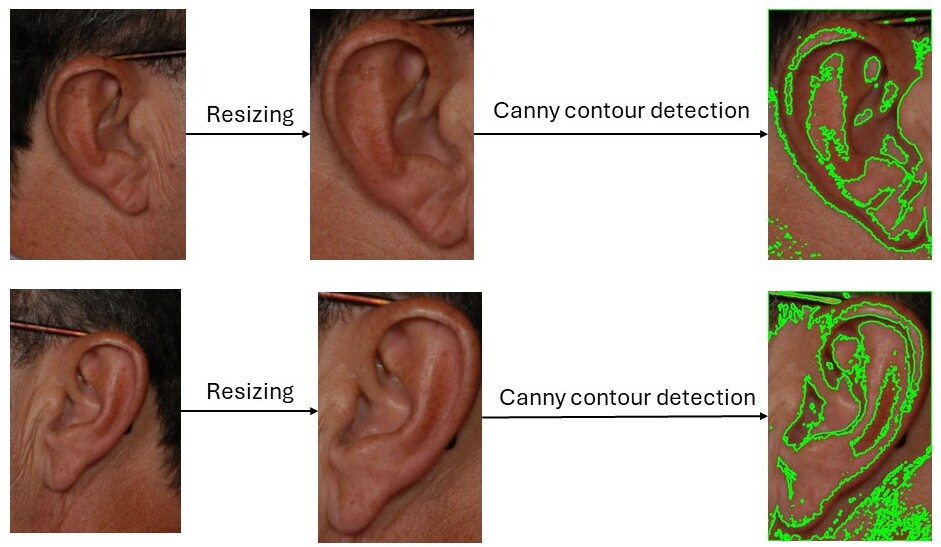}
    \caption{Ear biometric preprocessing stages.}
    \label{fig:image pipeline for preprocessing stages}
\end{figure}

Accurate ear localization and segmentation are crucial for ear biometric identification systems. This process relies heavily on a technique called contour detection \cite{28}, which extracts the outlines of ear structures from digital images. Essentially, contour detection isolates the ear region from the background and other facial features. Despite challenges presented by noise and low-contrast regions, contour detection remains an essential tool for ensuring accurate ear recognition in biometric systems. Canny edge detection algorithm is employed to extract edge features from ear images \cite{29,30}. It begins with Gaussian filtering to reduce noise, enhancing edge detection accuracy. Non-maximum suppression preserves local maxima in gradient direction, and double thresholding categorizes pixels into strong, weak, or non-edge based on gradient magnitude. Edge tracking by hysteresis ensures edge continuity. The resulting edge map visualizes detected edges, revealing ear structural features. In figure \ref{fig:image pipeline for preprocessing stages}, all stages for preprocessing will be observed.

\subsection{Augmentation} 
\begin{figure}[htpb]
    \centering 
    \includegraphics[width=0.4\textwidth]{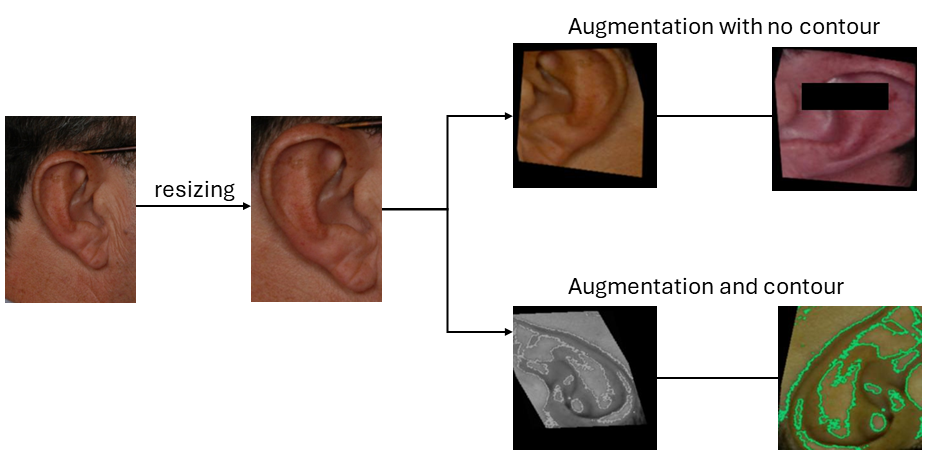}
    \caption{Ear Biometric Stages}
    \label{fig:image pipeline}
\end{figure}

Data augmentation enhances ear recognition models by introducing synthetic variations \cite{31}, expanding the dataset and aiding adaptation to real-world conditions. This reduces overfitting, improves generalization to unseen images, and enhances overall model performance. Geometric and photometric transformations are applied to enhance model robustness and improve performance. 
Geometric transformations, such as rotation, flipping, cropping, affine and perspective transformations, introduce variations in image orientation, scale, and perspective, simulating different viewing angles and real-world distortions. Photometric transformations manipulate color and grayscale properties, emulating changes in lighting conditions and testing model robustness to grayscale images. These transformations collectively enhance the model's ability to learn diverse features and improve its generalization capabilities. In Figure \ref{fig:image pipeline}, the overall pipeline from the initial stage of processing the image to its input into the model is depicted.

\section{Methodology}\label{sec:sample:sample6} 
For the ear biometric identification task, we utilize various convolutional neural network (CNN) architectures, including VGG16, VGG19, ResNet, MobileNet, and EfficientNet b7. In this section, we describe the architecture of each model.

\subsection{VGG16 \& VGG19} 
We employed the VGG-16 model as a pre-trained framework for ear identification, customizing it to our datasets of 100 and 164 images, respectively. The VGG-16 architecture, comprising 13 convolutional layers, 5 pooling layers, and 3 fully connected layers, processes input images of size $224 \times 224 \times 3$. We tailored the classifier section for our application. Additionally, we utilized the VGG-19 model, freezing the first four convolutional blocks and training the upper layers to refine high-level features, while customizing the final fully connected layer to accommodate the different number of classes in each dataset.

\subsection{ResNet50} 
ResNet-50, known for its depth and residual learning approach, was selected for its effectiveness in image classification tasks. The model is designed for \(224 \times 224 \times 3\) input images and comprises 16 residual blocks, each containing convolutional layers and shortcut connections to mitigate the vanishing gradient problem. To adapt ResNet-50 to our task, we employed a pretrained model and customized its classifier component by replacing the fully connected layer with a new linear layer tailored to our dataset's class requirements. 

\subsection{MobileNet v2}
MobileNet V2's lightweight design makes it suitable for resource-constrained devices without sacrificing performance utilizing depthwise separable convolutional layers. Transfer learning is employed by replacing the last fully connected layer and training only its parameters, while keeping the rest of the model's parameters fixed. 

\subsection{EfficientNet B7}
EfficientNet-B7's superior efficiency and scalability make it ideal for state-of-the-art performance in image classification tasks while conserving computational resources. The architecture, designed for \(224 \times 224 \times 3\) input images, comprises depthwise-separable convolutions, Swish activation functions, and squeeze-and-excitation blocks recalibrate channel-wise feature responses, enhancing discriminability. Transfer learning is applied by utilizing a pretrained model, freezing the front layers, and replacing the last linear layer of the classifier to match the classification requirements of our dataset.
\section{Experment and Result}\label{sec:sample:sample7}
\subsection{Experiments setup}
In this section, the experimental setup conducted for ear biometric identification using two distinct datasets: AMI and EarVN1.0, is delineated. Each experiment represents an independent stage aimed at enhancing the performance of the identification system. Specifically, all datasets were input into models including VGG16, VGG19, ResNet50, MobileNet, and EfficientNet B7 to evaluate their performance under various conditions. The following experiments were employed to explore different enhancement strategies:
\newline
1. \textbf{Baseline Model Evaluation(BM)} The datasets were directly inputted into the models without any preprocessing to establish a baseline performance as in Figure \ref{fig:five-images}.
\newline
2. \textbf{Preprocessing for Enhanced Detection (PP):} We applied image zooming and canny edge detection to improve ear detection accuracy.
\newline
3. \textbf{Augmentation and Zooming (AZ):} We applied augmentation to the zoomed ear images, increasing dataset diversity and robustness without incorporating previous preprocessing steps. Ten augmentations were applied to each image, Figure \ref{fig:image pipeline for preprocessing stages}.
\newline
4. \textbf{Comprehensive Enhancement Strategy (CES):} We employed a comprehensive strategy, combining image zooming, canny edge detection, and augmentation, Figure \ref{fig:image pipeline}. 

The experiments were conducted using a Kaggle notebook with a P100 GPU for training.
\subsection{Result}
In this section, we will discuss the results and observations that have been made for the two datasets, AMI and EarVN1.0, respectively.
\subsubsection{AMI Dataset}
\begin{figure}[t]
    \centering 
    \includegraphics[width=0.485\textwidth]{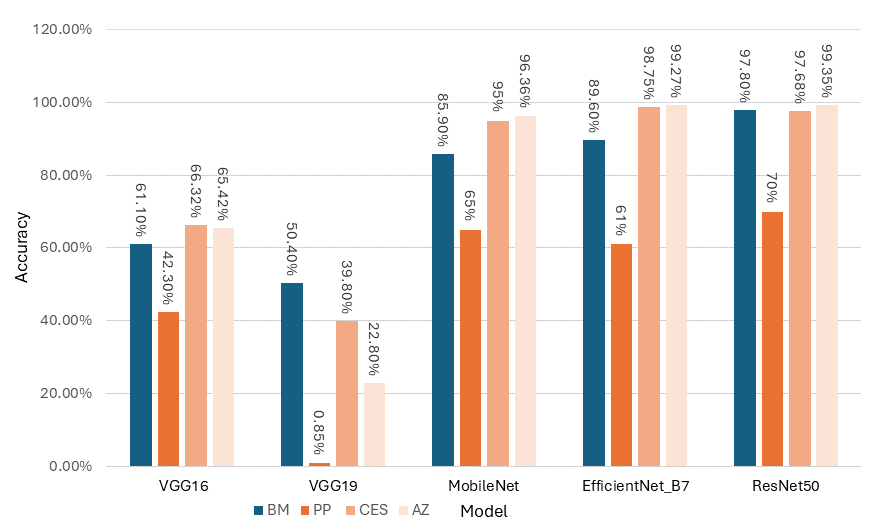}
    \caption{Comparison between Base Model (BM) vs. Preprocessing and Enhanced Detection (PP), Augmentation and Zooming (AZ), and Comprehensive Enhancement Strategy (CES) on the AMI dataset.}
    \label{fig:comparison4}
\end{figure}

Firstly, the BM was compared with all experiments conducted in this study. For CES, a 5\% improvement in testing accuracy was observed, while for AZ, testing accuracy also improved by 6\%. However, for PP, the testing accuracy experienced a 30\% worsening. Figure \ref{fig:comparison4} displays all experiments in comparison with BM.

\begin{table}[b]
\centering
\footnotesize
\caption{Experimental results for the AMI dataset.}
\label{tab:experimental-results}
\resizebox{0.485\textwidth}{!}{
    \begin{tabular}{|c|c|c|c|c|c|c|c|c|c|c|}
    \hline
    \textbf{Architecture} & \multicolumn{2}{c|}{\textbf{Base model}} & \multicolumn{2}{c|}{\textbf{Canny Edge}} & \multicolumn{2}{c|}{\textbf{Canny and Aug.}} & \multicolumn{2}{c|}{\textbf{Augmentation}} \\ \hline
     & \textbf{Train} & \textbf{Test} & \textbf{Train} & \textbf{Test} & \textbf{Train} & \textbf{Test} & \textbf{Train} & \textbf{Test} \\ \hline
    VGG16 & $86.7\%$ & $61.1\%$ & $88.8\%$ & $42.3\%$ & $99.3\%$ & $\textbf{\color{black}66.3\%}$ & $97.5\%$ & $65.4\%$ \\ \hline
    VGG19 & $81.0\%$ & $\textbf{\color{black}50.4\%}$ & $3.4\%$ & $0.85\%$ & $79\%$ & $39.8\%$ & $44.4\%$ & $22.8\%$ \\ \hline
    MobileNet & $100\%$ & $85.9\%$ & $100\%$ & $65.0\%$ & $100\%$ & $95\%$ & $100\%$ & $\textbf{\color{black}96.4\%}$ \\ \hline
    EfficientNet B7 & $100\%$ & $89.6\%$ & $100\%$ & $61.0\%$ & $100\%$ & $98.8\%$ & $99.9\%$ & $\textbf{\color{black}99.3\%}$ \\ \hline
    ResNet50 & $100\%$ & $97.8\%$ & $100\%$ & $70\%$ & $100\%$ & $98.8\%$ & $100\%$ & $\textbf{\color{black}99.4\%}$ \\ \hline
    \end{tabular}%
}
\end{table}

Secondly, using PP didn't improve our dataset compared to BM In fact, it led to a decline in our accuracy. However, when doing CES which is composed of contour and augmentation, there was an improvement observed, with training accuracy increasing by 11 - 75.6\% and testing accuracy increasing by 24 - 39\%. Finally, the highest accuracy achieved on the dataset was in the AZ, reaching an impressive 99.35\%. This stands in stark contrast to the second-best performance, which had an accuracy difference of only approximately 1\%. Table \ref{tab:experimental-results} illustrates the accuracies among all experiments and models for AMI Dataset.

\begin{figure}[t]
    \centering 
    \vspace*{\fill}
    \includegraphics[width=0.485\textwidth]{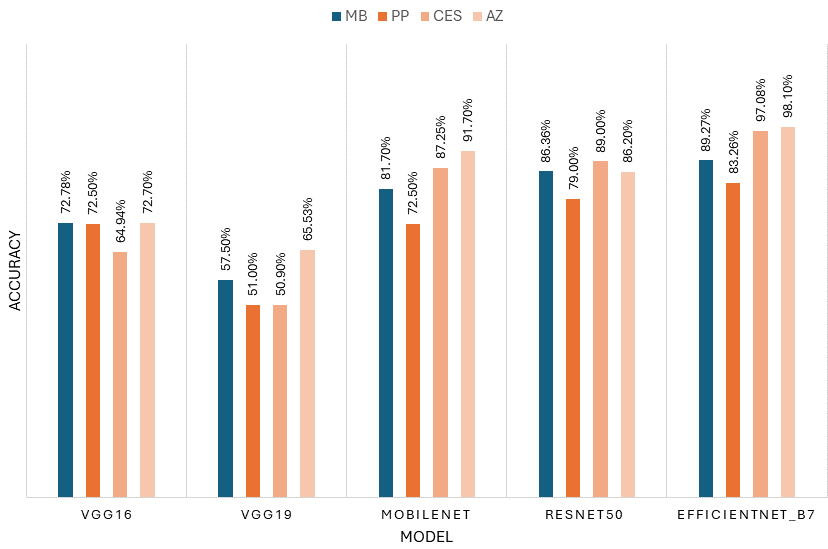}
    \caption{Comparison between Base Model (BM) vs. Preprocessing and Enhanced Detection (PP), Augmentation and Zooming (AZ), and Comprehensive Enhancement Strategy (CES) on the EarNV1.0 dataset.}
    \label{fig:comparisonn1}
\end{figure}

Comparison among different experiments revealed that CES outperformed other methods, showing a testing accuracy improvement of 6\%, while AZ exhibited a modest increase of 2\%. However, PP resulted in a decrease in accuracy. Specifically, in the AMI dataset, PP did not enhance testing accuracy compared to the baseline model (BM), whereas CES demonstrated an improvement of 10-15\%. This improvement can be attributed to augmentation, suggesting that contour detection did not contribute significantly to model enhancement. Consistently, AZ remained the most successful experiment with a testing accuracy of 98.1\%, followed closely by CES at 97.08\%. These findings suggest that contour detection negatively impacts model accuracy. (See Figure \ref{fig:comparisonn1} and Table \ref{tab:experimental-results} for details.)

\section{Discussion}
In this section, a comprehensive discussion of the findings and implications derived from our investigation into ear biometric identification leveraging deep learning methodologies is undertaken. 
\textbf{Datasets}: Accessing suitable datasets posed initial challenges for the research, highlighting the importance of comprehensive and diverse datasets for conducting thorough experiments and gaining meaningful insights. However, the datasets used, especially the AMI dataset, were limited in the number of images available. This constraint heightened the risk of overfitting, potentially leading the model to memorize the training data instead of learning meaningful patterns.

\begin{table}[b]
\centering
\footnotesize

\resizebox{0.485\textwidth}{!}{
    \begin{tabular}{|c|c|c|c|c|c|c|c|c|c|c|}
    \hline
    \textbf{Architecture} & \multicolumn{2}{c|}{\textbf{Base model}} & \multicolumn{2}{c|}{\textbf{Canny Edge}} & \multicolumn{2}{c|}{\textbf{Canny and Aug.}} & \multicolumn{2}{c|}{\textbf{Augmentation}} \\ \hline
     & \textbf{Train} & \textbf{Test} & \textbf{Train} & \textbf{Test} & \textbf{Train} & \textbf{Test} & \textbf{Train} & \textbf{Test} \\ \hline
    VGG16 & $99.4\%$ & $\textbf{\color{black}72.78\%}$ & $99.6\%$ & $72.5\%$ & $97.7\%$ & $65.0\%$ & $97.2\%$ & $72.7\%$ \\ \hline
    VGG19 & $92.4\%$ & $57.5\%$ & $94.3\%$ & $51.0\%$ & $88.3\%$ & $50.9\%$ & $97.2\%$ & $\textbf{\color{black}65.6\%}$ \\ \hline
    MobileNet & $99.8\%$ & $81.7\%$ & $99.6\%$ & $72.5\%$ & $97.9\%$ & $87.3\%$ & $96.9\%$ & $\textbf{\color{black}91.7\%}$ \\ \hline
    ResNet50 & $99.9\%$ & $86.4\%$ & $99.9\%$ & $79.0\%$ & $99.9\%$ & $\textbf{\color{black}89.0\%}$ & $93.5\%$ & $86.2\%$ \\ \hline
    EfficientNet B7 & $99.9\%$ & $89.3\%$ & $99.9\%$ & $83.3\%$ & $99.7\%$ & $97.1\%$ & $99.7\%$ & $\textbf{\color{black}98.1\%}$ \\ \hline
    \end{tabular}%
}
\caption{Experimental Results for EarNV1.0 dataset}
\label{tab:experimental-results2}
\end{table}

\textbf{Image Quality}: Another significant challenge arose from the quality of images within the datasets, notably the EarNV1.0 dataset. Poor image quality could hinder the training process and diminish the model's ability to accurately distinguish between ear features. Showing that at \ref{fig:samples-NV}
  
\begin{figure}[H]
  \centering
  \begin{tabular}{c@{\hspace{0.1cm}}c}
    \includegraphics[width=2.28cm]{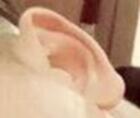} &
    \includegraphics[width=2cm]{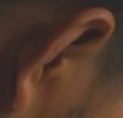 } \\
  \end{tabular}
  \caption{Examples of Pose variations for ear images}
  \label{fig:Pose challanging}
\end{figure}

\textbf{Pose Variation}: The orientation and pose of the ear relative to the camera can vary significantly across images, affecting the visibility and clarity of ear features. Dealing with pose variation requires techniques for pose normalization or robust feature extraction that can handle variations in ear orientation. Figure\ref{fig:Pose challanging} you can see some ear variation poses.

\textbf{Presence of Hair}: The presence of hair in ear images posed a significant obstacle during classification. Hair could obscure crucial ear features, leading the model to focus erroneously on non-relevant details and potentially compromising classification accuracy. \newline

Addressing these challenges is crucial for advancing the effectiveness and robustness of ear biometric identification systems based on deep learning methodologies. Through careful consideration and mitigation of these issues, the aim is to enhance the reliability and applicability of the classification models in real-world scenarios.

\section{Conclusion and Future works}
This paper introduces a novel approach to ear biometric identification, leveraging deep learning techniques to develop a robust and efficient system. We utilize ear images from different positions of the same individual to enhance recognition accuracy. We experiment with two datasets, AMI and EarNV1.0, comprising 700 and 28,412 images, respectively. To augment the data, we apply techniques such as random rotation, color jitter, and random resizing. Our results show that ResNet50 achieves 99.35\% testing accuracy for the AMI dataset, while EfficientNet B7 achieves 98.1\% accuracy for the EarNV1.0 dataset. Our proposed methods demonstrate performance enhancements, with an average improvement of 2\% for the AMI dataset and 1-4\% for the EarNV1.0 dataset across various models. These results underscore the effectiveness of our contributions in advancing ear biometric systems.\\
Future efforts should focus on acquiring larger, diverse datasets with high-quality images captured in varied environmental conditions. Developing more accurate and robust edge detection techniques tailored to ear biometric identification is also crucial. Investigating advanced algorithms for detecting and extracting ear contours with high precision will enhance the accuracy and reliability of the identification system.

\end{document}